%% file: main.tex
\RequirePackage{amsmath} %To get rid of the error "Unable to redefine math accent \vec."

\documentclass{llncs}

\usepackage{graphicx}
\usepackage{amssymb, amsfonts}
\usepackage{xfrac}
\usepackage[sectionbib,numbers]{natbib}
\usepackage{algorithm}
\usepackage[noend]{algorithmic}
\usepackage[titlenumbered,ruled,algo2e]{algorithm2e}
\usepackage{subfigure}
\usepackage{url}
\usepackage{ulem}
\usepackage{chngcntr}
\counterwithin{figure}{section}
\usepackage{float}
\usepackage{multirow}
\usepackage{color}\definecolor{darkblue}{rgb}{0,0,0.75}
\usepackage[pdftex,pdftitle="Resource-Constrained Federated Learning",colorlinks=true, bookmarksnumbered, citecolor=darkblue, urlcolor=darkblue, linkcolor=darkblue, pdfpagemode=UseNone]{hyperref}
\usepackage{wrapfig}
\input{preamble}
\usepackage{booktabs}

\title{Resource-Constrained On-Device Learning by Dynamic Averaging}
\date{}
\author{
Lukas Heppe\inst{1}\and
Michael Kamp\inst{2}\and
Linara Adilova\inst{3,4}\and
Danny Heinrich\inst{1}\and
Nico Piatkowski\inst{3}\and
Katharina Morik\inst{1}
}
\institute{
TU Dortmund \email{<name>.<surname>@tu-dortmund.de}
\and
Monash University \email{michael.kamp@monash.edu}
\and
Fraunhofer IAIS \email{<name>.<surname>@iais.fraunhofer.de}
\and
Fraunhofer Center for Machine Learning
}

\begin{document}

\maketitle

\begin{abstract}
The communication between data-generating devices is partially responsible for a growing portion of the world's power consumption. Thus reducing communication is vital, both, from an economical and an ecological perspective. For machine learning, on-device learning avoids sending raw data, which can reduce communication substantially. Furthermore, not centralizing the data protects privacy-sensitive data. However, most learning algorithms require hardware with high computation power and thus high energy consumption. In contrast, ultra-low-power processors, like FPGAs or micro-controllers, allow for energy-efficient learning of local models. Combined with communication-efficient distributed learning strategies, this reduces the overall energy consumption and enables applications that were yet impossible due to limited energy on local devices. The major challenge is then, that the low-power processors typically only have integer processing capabilities. 
This paper investigates an approach to communication-efficient on-device learning of integer exponential families that can be executed on low-power processors, is privacy-preserving, and effectively minimizes communication. The empirical evaluation shows that the approach can reach a model quality comparable to a centrally learned regular model with an order of magnitude less communication. Comparing the overall energy consumption, this reduces the required energy for solving the machine learning task by a significant amount.
\end{abstract}

\input{introduction}

\input{theory}

\input{experiments}

\input{conclusion}

\bibliographystyle{splncsnat}
\bibliography{bibliography}

\end{document}

%% file: preamble.tex
\newtheorem{pr}{Proposition}

\newtheorem{cor}[pr]{Corollary}

\newsavebox\IBoxA \newsavebox\IBoxB \newlength\IHeight
\newcommand\TwoFig[6]{% Image1 Caption1 Label1 Im2 Cap2 Lab2
  \sbox\IBoxA{\includegraphics[width=0.48\textwidth]{#1}}
  \sbox\IBoxB{\includegraphics[width=0.48\textwidth]{#4}}%
  \ifdim\ht\IBoxA>\ht\IBoxB
    \setlength\IHeight{\ht\IBoxB}%
  \else\setlength\IHeight{\ht\IBoxA}\fi
  \begin{figure}[!htb]
  \minipage[t]{0.48\textwidth}\centering
  \includegraphics[height=\IHeight]{#1}
  \caption{#2}\label{#3}
  \endminipage\hfill
  \minipage[t]{0.48\textwidth}\centering
  \includegraphics[height=\IHeight]{#4}
  \caption{#5}\label{#6}
  \endminipage 
  \end{figure}%
}

\newcommand{\zv}[1]{\ensuremath{\boldsymbol{#1}}}
\newcommand{\dom}[1]{\ensuremath{\mathcal{#1}}}
\newcommand{\prob}{\ensuremath{\mathbb{P}}}

\newcommand{\ci}{\ensuremath{\perp\kern-5pt\perp}}

\newcommand{\transpose}{\ensuremath{^{\top}}}
\newcommand{\maxcliques}{\mathcal{C}}

\newcommand{\muhat}{\hat{\vec{\mu}}}

\newcommand{\defemph}[1]{\textbf{#1}}

\newcommand{\card}[1]{\left|#1\right|}
\newcommand{\abs}[1]{\left|#1\right|}

\newcommand{\R}{\mathbb{R}}
\newcommand{\N}{\mathbb{N}}
\newcommand{\Z}{\mathbb{Z}}
\newcommand{\expected}{\mathbb{E}}

\newcommand{\X}{\mathcal{X}}
\newcommand{\Y}{\mathcal{Y}}
\newcommand{\D}{\mathcal{D}}

\newcommand{\inner}[2]{\left\langle #1, #2 \right\rangle}

\newcommand{\algo}{\mathcal{A}}

\newcommand{\balancingSet}{\mathcal{B}}

%% file: introduction.tex
\section{Introduction}

Today more and more data is generated by physically distributed sources, e.g., smartphones, sensors, and IoT devices. Performing machine learning on this data not only poses severe challenges on the bandwidth, on storing and processing it, but also requires enormous amounts of energy: currently, the world communicates around $20\cdot 10^{10}$ GB per month~\citep{hootsuite2019digital} with a power consumption of around $0.3$kWh/GB~\citep{pihkola2018evaluating}, resulting in a total energy consumption of $6$TWh per month. In comparison, the largest nuclear plant in the US, the R.E. Gina reactor in Arizona, generates around $0.39$TWh per month~\citep{american2018how}, so more than $15$ such reactors are needed just to power the communication. With an estimated number of $5\cdot 10^{10}$ connected devices by the end of 2020~\citep{mohan2016edge}, the amount of communication will grow substantially, with some of them (e.g., autonomous vehicles or airplanes) communicating up to $5$GB per second~\citep{shi2016edge}. Thus, machine learning on this data could become responsible for a large portion of the world's power consumption.

In order to reduce the communication load, models can be trained locally and only model parameters are centralized periodically~\citep{mcmahan2017communication} or dynamically~\citep{kamp2018efficient, kamp2016communication}. However, this approach requires sufficient computation power at the local data sources - this is usually available on smartphones, but not necessarily on sensors or IoT devices. Most sensors or IoT devices could be fit with efficient, low-power processors which typically only have integer processing capabilities - no floating point unit. Recently, it was shown that learning exponential families can be performed on such devices using only integer and bit operations~\citep{piatkowski2018exponential}. Scaling this learning to the internet of things requires to implement communication-efficient distributed learning on these devices, too. 

This work proposes a resource- and communication-efficient distributed learning approach for exponential families that uses only integer and bit operations. The key idea is to only communicate between local devices if their model is sufficiently different from the global mean, implying that it has learned truly novel information. Hence, we adapt dynamic averaging~\citep{kamp2014communication} to require only integer operations. In this approach, each device checks a sufficient local condition and only communicates if it is violated. In case of a violation, model parameters are centralized and averaged into a joint model, which is redistributed to the devices. Setting the maximum allowed divergence between a model and the global mean allows users to the trade-off between communication-efficiency and joint model quality.

We theoretically analyze this approach and provide guarantees on the loss and bounds on the amount of communication. We show empirically that using resource-constrained dynamic averaging on integer exponential families allows to reach a model performance close to full floating point models and reduces the required communication substantially.

\subsubsection{Related work} 
Several works have been published, minimizing the resource-consumptions in federated learning environments. However, most publications already consider smartphones as resource-constrained environments, while we go one step further and focus on ultra-low-power hardware without access to floating-point-units. A similar work has been done by Piatkowski \cite{Piatkowski/2019a}, who also considered resource-constrained family models for a distributed learning task. Albeit, the reduced resource-consumption is based on parameter-sparsification and one-shot averaging. Besides, the model aggregates reside in regular model space and thus cannot be applied on ultra-low-power hardware. Alternative methods, proposed by Wang et. al., focus on selecting a trade-off between global aggregation and local updates, while we follow a distinct approach based on dynamic conditions. A survey on federated learning concepts and the challenges can be found in \cite{DBLP:journals/corr/abs-1909-11875}.

%% file: theory.tex
\section{Resource-Constrained Exponential Family Models}

Probabilistic graphical models form a subset of machine learning and combine graph with probability theory \cite{Wainwright/etal/2008a}. They are used to model complex probability distributions, which can be utilized to solve a variety of tasks. The following subsections will introduce the notation and background of graphical models in general as well as highlight the specifics for resource-constrained models.

\subsection{Undirected graphical models}

% Which variable for 

% Connection between nodes and variables

Let $G = (V,E)$ be an undirected graph with $|V| = n$ vertices which are connected by the edges $(s,t) \in E \subseteq V \times V$. A clique $C \subseteq V$ is formed by some fully connected subset of the vertices: $\forall\; u,v \in C, u \neq v : (u,v) \in E$. The maximal cliques of the graph are those, who are not contained in any larger clique. We denote the set of all maximal cliques by $\mathcal{C}$. In addition, let $\zv{X} = (\zv{X}_1, \hdots, \zv{X}_p)\transpose $ denote some random vector where each variable $\zv{X}_i$ can take values of a discrete set $\dom{X}_i$. In turn the vector can take values of the cross product from each variable $\dom{X} = \dom{X}_1 \times \hdots \times \dom{X}_p$. Besides, we allow the indexing of $\zv{X}$ and $\dom{X}$ by subsets like cliques, e.g., $\zv{X}_C$ and $\dom{X}_C$. Specific assignments $\vec{x}_C$ to those random variables are denoted by bold lowercase letters.

An undirected graphical model represents the joint distribution of $\zv{X}$ by exploiting conditional independencies between the variables. We model those using a graph, if $ (s,t) \notin E \implies \zv{X}_s \ci\zv{X}_t \; | \; \zv{X} \setminus \{ \zv{X}_s, \zv{X}_t  \}$. Thus, the graph encodes the conditional independence structure of the distribution. If we introduce potential functions $\psi_C : \dom{X}_C \mapsto \R_{+} $ for each maximal clique of the graph, then according to the Hammersley-Clifford-theorem \cite{HammersleyClifford1971theorem} the density factorizes as follows
\begin{equation}
\label{eq:joint-prob}
\prob(\zv{X} = \vec{x}) = \dfrac{1}{Z} \prod_{C \in \mathcal{C}} \psi_{C}(\vec{x}_{C}) \enspace,
\end{equation}
where $Z = \sum_{ \vec{x} \in \dom{X}} \prod_{C \in \mathcal{C}} \psi_{C}(\vec{x}_{C}) $ acts as normalizer in order to ensure that $\prob$ is a valid probability distribution. By utilizing probabilistic inference algorithms like belief propagation, this distribution can be used for a variety of tasks like querying marginal- or conditional probabilities as well as computing (conditional) maximum-a-posteriori estimates, which, in turn, can be used for predictive tasks.

\subsection{From regular to resource-constrained models}

% basically, those phi transform assignments to the variables of the cliques to one hot-encoded vectors

Let $C \in \maxcliques$ be a maximal clique of $G$, $\vec{\theta}_C \in \R^{\abs{\dom{X}_C}}$ a parameter vector and $\phi : \dom{X}_C \mapsto \{0, 1\}^{\abs{\dom{X}_C}} $ some feature function or sufficient statistic, which maps assignments of the variables to one-hot-encoded vectors. By defining $ \psi_C (\vec{x}_C) = \exp{ \inner {\vec{\theta}_C}{\phi_C(\vec{x}_C}) }$, concatenating all $\vec{\theta}_C$ in $\vec{\theta}$ as well as all $\phi_C$ in $\phi$, the joint distribution can be represented as the canonical exponential family \cite{Wainwright/etal/2008a} 
\begin{equation}
    \label{eq:exponential-family}
    \prob_{\vec{\theta}} (\zv{X} = \vec{x}) = \exp({\inner{\vec{\theta}}{\phi(\vec{x})} - A(\vec{\theta})}) \enspace,
\end{equation}
with $A(\vec{\theta}) = \log Z(\vec{\theta})$, which contains many well known distributions like the gamma, normal, and exponential distribution. Despite its compactness, in contrast to other models like neural networks with millions of parameters, the model cannot be evaluated on ultra low power devices since it requires the presence of floating point units. 

Driven by the cheap availability of ultra low power hardware, the increased power consumption of machine learning models and the trend to push those towards the edge, Piatkowski \cite{DBLP:journals/ijon/Piatkowski0M16} developed a variation of the regular exponential family, which is capable of running on devices, that only have access to integer arithmetic units. By replacing the base in equation \ref{eq:exponential-family} from $e$ to $2$ and restricting the parameters to be a subset of $\N_{\le k}$ with fixed word size $k$, the resource constrained exponential family is defined as follows
\[
	P_{\vec{\theta}}(\mathbf{X} = \vec{x} )	= 2^{\inner{\vec{\theta}}{\phi(\vec{x})} - A(\vec{\theta})}
\]
with
\[
	A(\vec{\theta}) = \log_{2} \sum_{\vec{x} \in \mathcal{X}} 2^{\inner{\vec{\theta}}{\phi(\vec{x})} - A(\vec{\theta})  }\enspace.
\]
The availability of specialized integer versions of the inference- and optimization-algorithms does allow not only the application, but also the learning of these models directly on local devices. Moreover, Piatkowski has proven \cite{piatkowski2018exponential}, that despite the limitations the models still provide theoretical guarantees on the quality.

\subsubsection{Learning}

The parameters  $\vec{\theta}$ of the distribution $\prob_{\vec{\theta}}$ are estimated using a (regularized) maximum likelihood estimation. Suppose we are given a dataset $\mathcal{D} = \{ \vec{x}_{1},\hdots,\vec{x}_n\}$ with $n$ samples, the negative average log-likelihood is defined as follows
\[
\ell(\vec{\theta}; \D) =  \log_{2} A(\vec{\theta}) - \inner{\vec{\theta}}{\dfrac{1}{\abs{\D}} \sum_{\vec{x} \in \D} \phi(\vec{x}) }
\]
Setting $\hat{\vec{\mu}} = \dfrac{1}{\abs{\D}} \sum_{ \vec{x} \in \D } \phi(\vec{x} )$ the partial derivative of $\ell$ is as follows
\[
\dfrac{\partial \ell(\vec{\theta}; \D)}{\partial \vec{\theta}_i} = \expected_{P} [ \phi(\vec{x})_i ] - \hat{\vec{\mu}}_{i} \enspace,
\]
which is just the difference between the empirical and the model's distribution. The model's distribution is computed using the BitLength-Propagation-algorithm \cite{piatkowski2018exponential}, which returns the probabilities as quotients $a / b$ to avoid floating point numbers. Likewise for $\hat{\vec{\mu}}$, we store the raw counts as well as the cardinality of the dataset as integers. Using a proximal block coordinate descent method, the parameters for each clique are updated by either increasing or decreasing the current value by one. In-depth details on those specialized integer algorithms can be found in \cite{piatkowski2018exponential}. If the samples $\vec{x}$ arrive as stream, e.g., sensor readings, the $\phi$'s and example's counters will be accumulated iteratively and used for a gradient step. This is showcased in algorithm \ref{alg:int-update}.

\begin{algorithm}[t]
	\caption{Integer Learning Algorithm}
	\label{alg:int-update}
	\textbf{Input:} Graph $G =(V,E)$, bytes $k \in \N$\\ % Batchsize??   
	\textbf{Initialization:}\\
	\vspace{-0.4cm}
	\begin{algorithmic}[0]
		\STATE local sufficient statistics $\muhat_{1}^{1},\hdots,\muhat_{1}^{m} \leftarrow \vec{0}$
		\STATE local example counter $ \enspace  \enspace c_{1}^{1},\hdots,c_{1}^{m}  \enspace \leftarrow 0 $ 
%		\STATE reference vector $r \leftarrow \theta$
%		\STATE violation counter $v \leftarrow 0$
	\end{algorithmic}
	\textbf{Round }$t$\textbf{ at learner }$i$\textbf{:}\\
	\vspace{-0.4cm}
	\begin{algorithmic}[0]
		\STATE \textbf{observe} $S_t^i\subset\X$
		\FOR{$\vec{x} \in S_t^i$}
		\STATE $\muhat_{t}^{i} \leftarrow \muhat_{t}^{i} + \phi(\vec{x}) $
		\ENDFOR
		\STATE $c_{t}^{i} \leftarrow c_{t}^{i} + \abs{S_t^i}$
		\STATE $\vec{\mu}_{t}^{i} \leftarrow$ \texttt{BitLengthBP()}  \tcp*{Compute models distribution}
		\STATE $\nabla \leftarrow \texttt{Grad(}\vec{\mu}_{t}^{i}\texttt{,}\vec{\mu}_{t}^{i}, c_{t}^{i} )$  \tcp*{Compute grad using Int-Prox}
		\STATE $\vec{\theta}_{t}^{i} \leftarrow \texttt{ApplyGrad}(\vec{\theta}_{t-1}^{i}, \nabla)$
		\STATE \textbf{return} $\vec{\theta}_{t}^{i}$
		
	%	\STATE \textbf{update} $\emphi^{i}_{t} = \emphi^{i}_{t-1} + \phi(S_t^i)  $
	%	\STATE \textbf{update} $\theta^i_{t-1}$ using the learning algorithm $\algo$\\

	\end{algorithmic}
\end{algorithm}

\section{Distributed Learning of Integer Exponential Families}
The integer exponential families described above can be trained on a data-generating device using only integer computations. The goal of distributed learning is to jointly train a model across multiple devices. That is, we assume a set of $m\in\N$ local devices, denoted \defemph{learners}, learning a joint task defined by a target distribution $\D:\X\times\Y\mapsto\R_+$. The learners obtain local samples over time. For simplicity, we assume rounds $t=1,2,\dots$ where in each round, each learner $l\in [m]$ obtains a local dataset $S^l_t\subset\X\times\Y$ drawn iid. from $\D$. 

The most straight-forward approach to solve this task is to compute local data summaries $\widehat{\vec{\mu}}^l_t$ on all observed data $\bigcup_{i=1}^t S^l_i$ as
\[
\widehat{\vec{\mu}}^l_t = \frac{1}{t}\sum_{i=1}^t\frac{1}{\left|S^l_i\right|}\sum_{ \vec{x} \in S^l_i}\phi(\vec{x})\enspace .
\]
%TODO: Should we define n = \sum_{i=1}^{m} m_l and n_l = \frac{n_l}/{n}??
These data summaries can then be centralized and the global data summary $\widehat{\vec{\mu}}$ is computed as the weighted average, where $n_l$ is the number of samples the learner $l$ received and $n=\sum_{i=1}^{l} n_l$ is total amount of samples accros all learners
\begin{equation}
\widehat{\vec{\mu}}_t=\sum_{l=1}^m \frac{n_l}{n} \widehat{\vec{\mu}}^l_t\enspace .
\end{equation}
With this data summary, the respective $\vec{\theta}_t$ can be computed centrally and thus we call this the \defemph{centralized} approach.
Since $\widehat{\vec{\mu}}_t$ is the exact data summary of the union of local dataset, this approach results in the same model as learning on the union of all local datasets, directly. However, it has two major disadvantages: it does not make use of the local computing power at the data-generating devices and it requires centralizing potentially sensitive data.

To overcome these disadvantages, we propose to train models locally to obtain both $\vec{\theta}^l$ and $\vec{\mu}^l$ for each learner $l\in [m]$. We synchronize these local models by averaging the parameters. The average of a set of integer vectors, however, is not necessarily an integer. Instead, the floored average can be computed using only integer operations~\footnote{Indeed, the average of two integers in binary representation can be computed using only the logical ``and'' $\&$ and ``or'' $+$ operations, as well as the bit-shift operator ``$>>$'' as $\left\lfloor\frac{a+b}{2}\right\rfloor = (a\enspace\&\enspace b) + \left((a\enspace XOR\enspace b) >> 1\right)$.}. 
Of course, averaging both, $\vec{\theta}$ and $\vec{\mu}$, does not solve the privacy issue, since $\vec{\mu}$ is shared just like in the centralized approach. Thus, we refer to this as \defemph{na\"ive averaging} and use it only as a baseline. 
Instead, we propose to only average the model parameters $\vec{\theta}$ and maintain local data summaries $\vec{\mu}$. We call this approach \defemph{privacy-preserving resource constrained averaging}. 
\begin{table}[t]
    \centering
    \caption{Summary of transmitted data}
\begin{tabular}{c|c|c|c}
	\toprule
	Protocol & Centralized & Na\"ive & Privacy   \\ 
	\midrule 
	\midrule 
	Send & $\vec{\mu}_{i}^{t}$ & $\vec{\mu}_{i}^{t}$ + $\vec{\theta}_{i}^{t}$  & $\vec{\theta}_{i}^{t}$ \\ 
	\midrule 
	Receive & $\vec{\theta}_{Global}^{t}$ & $\widehat{\vec{\mu}_{i}^{t}}$ + $\widehat{\vec{\theta}_{i}^{t}}$    & $\widehat{\vec{\theta}_{i}^{t}}$ \\ 
	\bottomrule 
\end{tabular}

    \label{tab:data-in-protocol}
\end{table}

This averaging can be performed periodically, i.e., after observing $b\in\N$ batches, hence we call this \defemph{periodic averaging}. The frequency of averaging allows to balance communication and model quality: communication effort can be saved by averaging less frequently at the expense of model quality.

Communication can be further reduced by deciding in a data-driven way when averaging has the largest impact. \defemph{Dynamic averaging}~\citep{kamp2018efficient, kamp2014communication} checks local conditions to determine when to communicate. The algorithm is presented in Algo.~\ref{alg:protocol}. It shows the local computation at each round using the (integer) or real-value learning and the local test of the conditions.
That is, with a common reference point $\vec{r}\in\Z^d$, each local learner checks its local condition
\[
\|\vec{\theta}^i - \vec{r}\|_2^2 = \sum_{j=1}^d\left(\vec{\theta}^i_j - \vec{r}\right)^2 \leq \Delta\enspace ,
\]
where $\Delta\in\Z$ is a predefined threshold. 
% Delta is not Z^d?
If all model parameters $\vec{\theta}^1,\dots,\vec{\theta}^m\in\Z^d$ and the reference point $\vec{r}\in\Z^d$ are $d$-dimensional integer vectors, and the divergence threshold $\Delta\in\Z$ is also an integer, then the local conditions can be checked using only integer operations. 

The algorithm also shows the coordinator processing the submitted models. The augmentation requests additional parameter vectors in case of the dynamic averaging. The number of models received is doubled until the conditions are fulfilled. Hence, it may happen, that finally all learners are used for the central processing which gives a new global model to all local learners.

As mentioned before, using integer exponential families introduces an error into the models. Similarly, using rounded averaging introduces an error. This error can be bounded. Training regular exponential families is a convex learning problem. Indeed, it is trivial to show that the error of using $\widehat{\vec{\theta}}$ instead of the standard average $\overline{\vec{\theta}}$, i.e., $\|\widehat{\vec{\theta}} - \overline{\vec{\theta}}\|_2$ is bounded by $\sqrt{d}$, where $d$ denotes the number of parameters. Let $\epsilon\in\R_+$ denote a bound on the error of using an integer exponential family instead of a real-valued one. Furthermore, define the cumulative loss on $m\in\N$ learners until time $t\in\N$ as
\[
L(t,m) = \sum_{i=1}^t\sum_{l=1}^m\sum_{(\vec{x},y)\in S^l_t}\ell(\vec{x},y)\enspace ,
\]
where $\ell:\Y\times\Y\rightarrow\R_+$ is a loss function. Then, it follows directly from Cor.~3.33 in~\citet{kamp2019black} that when using stochastic gradient descent to train the local models, the cumulative loss of using resource-constrained dynamic averaging over normal periodic averaging is bounded. 
\begin{cor}
Assume $m\in\N$ learners jointly training an integer exponential family with stochastic gradient descent with learning rate $\eta\in\R_+$. Furthermore, assume there exists $\rho\in\R_+$ such that $\|x\|_2\leq\rho$ for all $x\in\X$. Let $L_{Per}, L_{Dyn}$ denote the cumulative loss when local models are maintained by resource-constrained periodic, resp. dynamic averaging. Then it holds that
\[
L_{Dyn}(t,m) - L_{Per}(t,m) \leq \frac{t}{b\eta^2\frac{\rho}{\rho^2 + 1}}(\Delta + 2d + \epsilon)\enspace .
\]
\end{cor}
Using the observation that for stochastic gradient descent, periodic averaging on $m\in\N$ learners with batch size $b=1$ is equal to centralized mini-batch SGD with mini-batch size $m$ and learning rate $\eta/m$ \citep[cf.][Prop. 3]{kamp2018efficient}, as well as using the standard learning rate of $\eta=\sqrt{t}$ and $\Delta=\sqrt{t}$, it follows that the regret of using resource-constrained dynamic averaging over centralized training is a constant in $d$ and $\epsilon$. 
In the following section, we empirically compare dynamic averaging to periodic averaging, as well as the centralized approach, both in terms of model quality and communication demand.

\begin{algorithm}[ht]
	\caption{Resource-Constrained Dynamic Averaging Protocol}
	\label{alg:protocol}
	\textbf{Input:} learning algorithm $\algo$, divergence threshold $\Delta\in\N$, parameter $b\in\N$, $m$ learners\\    
	\textbf{Initialization:}\\
	\vspace{-0.4cm}
	\begin{algorithmic}[0]
		\STATE local models $\theta^1_1,\dots,\theta^m_1 \leftarrow$ one random $\theta$
		\STATE reference vector $r \leftarrow \theta$
		\STATE violation counter $v \leftarrow 0$
	\end{algorithmic}
	\textbf{Round }$t$\textbf{ at learner }$i$\textbf{:}\\
	\vspace{-0.4cm}
	\begin{algorithmic}[0]
		\STATE \textbf{observe} $S_t^i\subset\X\times\Y$
		\STATE \textbf{update} $\theta^i_{t-1}$ using the learning algorithm $\algo$\\
		\IF{$t \mod b=0$ \textbf{and} $\|\theta^i_{t}-r\|_2^2> \Delta$}
		\STATE \textbf{send} $\theta^i_{t}$ to coordinator (violation)
		\ENDIF
	\end{algorithmic}
	%\bigskip
	\textbf{At coordinator on violation:}\\
	\vspace{-0.4cm}
	\begin{algorithmic}[0]
		\STATE \textbf{let} $\balancingSet$ be the set of learners with violation
		\STATE $v\leftarrow v+\card{\balancingSet}$
		\STATE \textbf{if} $v=m$ \textbf{then} $\balancingSet\leftarrow [m]$, $v\leftarrow 0$
		\WHILE{$\balancingSet \neq [m]$ \textbf{and} $\left\|\left\lfloor\frac{1}{\balancingSet}\sum_{i \in \balancingSet}\theta^i_t\right\rfloor-r\right\|^2> \Delta$}
		\STATE \textbf{augment} $\balancingSet$ by augmentation strategy
		\STATE \textbf{receive} models from learners added to $\balancingSet$
		\ENDWHILE
		\STATE \textbf{send} model $\widehat{\theta}=\left\lfloor\frac{1}{\balancingSet}\sum_{i \in \balancingSet}\theta^i_t\right\rfloor$ to learners in $\balancingSet$
		\STATE \textbf{if} $\balancingSet=[m]$ also set new reference vector $r\leftarrow\widehat{\theta}$
	\end{algorithmic}
\end{algorithm}

%% file: experiments.tex
\section{Experiments}

In our experiments, we want to empirically investigate the centralized, the public, and the private scheme of communication and distributed learning. We compare the periodic and the dynamic update in the distributed learning setting for both, the regular and the resource-constrained averaging operator. 
Before we introduce the specific research questions and results, we describe the experimental setup. To emphasize reproducible research, we selected two open-source-frameworks for our implementation and experiments. To simulate a distributed learning environment, we utilized the \texttt{Distributed-Learning-Platform}\footnote{\url{https://github.com/fraunhofer-iais/dlplatform}}. The model implementation was based on the \texttt{Randomfields}-library \footnote{\url{https://randomfields.org/}}. Also, the source code for the new aggregation-operator and experiments can be found online\footnote{\url{https://bitbucket.org/zagazao/dynamic-rc-averaging/src/master/}}. 

%TODO: Should we mention handling of overflows(it's done by randomfields-library)?

\subsection{Model Quality and Communication}
During the simulation of the distributed learning environment, we limited ourselves to $m=16$ learners, however, in future work we want to investigate, how this approach scales with an increasing number of learners. For each integer learner, we limited the number of bits for each parameter to $k=3$, which results in $2^3$ possible choices. Lower choices of $k$ resulted in significantly worse performance in comparison to regular exponential family models. Increasing the number of bits did not result in a sufficient increase in performance and had the disadvantage of higher memory- and thus communication complexity. For the evaluation, we chose three different datasets -- \texttt{DOTA2}, \texttt{COVERTYPE} and \texttt{SUSY} -- of the UCI-Repository \cite{Dua:2019}. The datasets possess different properties. While the \texttt{DOTA2}-dataset features mostly discrete columns, has many features but a low amount of samples, the \texttt{SUSY} dataset consists of only a few real-valued columns, while consisting of many samples. The \texttt{COVERTYPE}-dataset sits in between those two with a mix of discrete and numerical features and is a medium dataset size. Details on those datasets can be found in table \ref{tab:dataset-statistics}.

\begin{table}
\centering
\caption{Dataset and model properties}
\begin{tabular}{c|c|c|c|c|c|c}
	\toprule
	Dataset   & Samples & Features & Classes & Discrete & Numerical & Model Dimension\\ 
	\midrule 
	\midrule 
	\texttt{SUSY}      & 5.000.000 & 19 & 2 & 1 & 18 & 1620 \\ 
	
	\texttt{COVERTYPE} & 581.012 &  55 & 7 & 45 & 10 & 1596 \\ 
	 
	\texttt{DOTA2}     & 102.944 & 117 & 2 & 117 & 0 & 2790 \\ 
	\bottomrule 
\end{tabular}
\label{tab:dataset-statistics}
\end{table}

All datasets have undergone the same preprocessing: numerical columns have been discretized into ten bins based on their quantiles. Furthermore, a random subset of 10.000 examples was selected as a holdout set, which in turn was used to estimate the model's structure via the Chow-Liu-algorithm \cite{Chow/Liu/1968a}. This step has to be done in order to ensure that all models share the same structure. Otherwise, the aggregation would not be possible. Note that this a serious limitation and future work could also investigate if aggregation can be applied to nodes with distinct graph structures. The remaining data was partitioned horizontally alongside the nodes.

% for a specified number of iterations $o$, defining an optimization-budget 
During the running-phase, at each time step $t$ all learners received a batch of $bs=10$ new samples. As soon as the batch arrived, learners are asked to predict the labels for each of the samples. Afterwards, the learner uses the new samples in order to update its local data summary $\widehat{\vec{\mu}}$. Thereafter, except for the centralized approach, where a global model is fitted on the accumulated data summaries, learners run an optimization-algorithm with the current data summaries and weights for a specified optimization-budget $o$ in terms of iterations. The budget allows for a further trade-off between model-quality and battery life of some device. Finally, each learner checks if synchronization should be performed. In case of the periodic protocol, the model parameters and/or data summaries are transmitted to the central coordinator after $b$ batches have been processed. In contrast, the dynamic protocol checks if its local conditions hold and only communicates if some condition is violated.

Specifically we want to answer the following {\bf questions}:
 \begin{itemize}
  \item How does the periodic protocol compare against the dynamic one?
  \item How do resource-constrained models compare to regular ones in terms of predictive quality? %Private RC and Normal
  %\item What are the particular well-suited intervals of updates for a trade-off regarding accuracy and communication efforts?
  %How do the different aggregations schemes compare to each other in terms of communication and model quality?
 \end{itemize}
 
\noindent
We compare the methods against {\bf two baselines}:
\begin{itemize}
    \item  {\bf Global Learning} We want to compare ourselves against the traditional machine learning setting, where all the data is centralized first. The performance of the global model, a regular exponential family, is evaluated using a 5-fold cross-validation procedure.  In the plots this is denoted by \texttt{Global}.
%TODO: Now I want to have to change the text:D    

%    In this setting we used the centralized averaging scheme with a regular MRF and a period of $1$. Distributed models should reach a similar performance to this one.
 \item {\bf Local learning} In this setting each device fits a model based on its local data. No communication takes place, thus this setting acts as a guard, showing that communication helps to improve model quality. In the plots this is denoted by \texttt{NoSync}.
\end{itemize}

\begin{figure}[t]
    \centering
    \includegraphics[width=\linewidth]{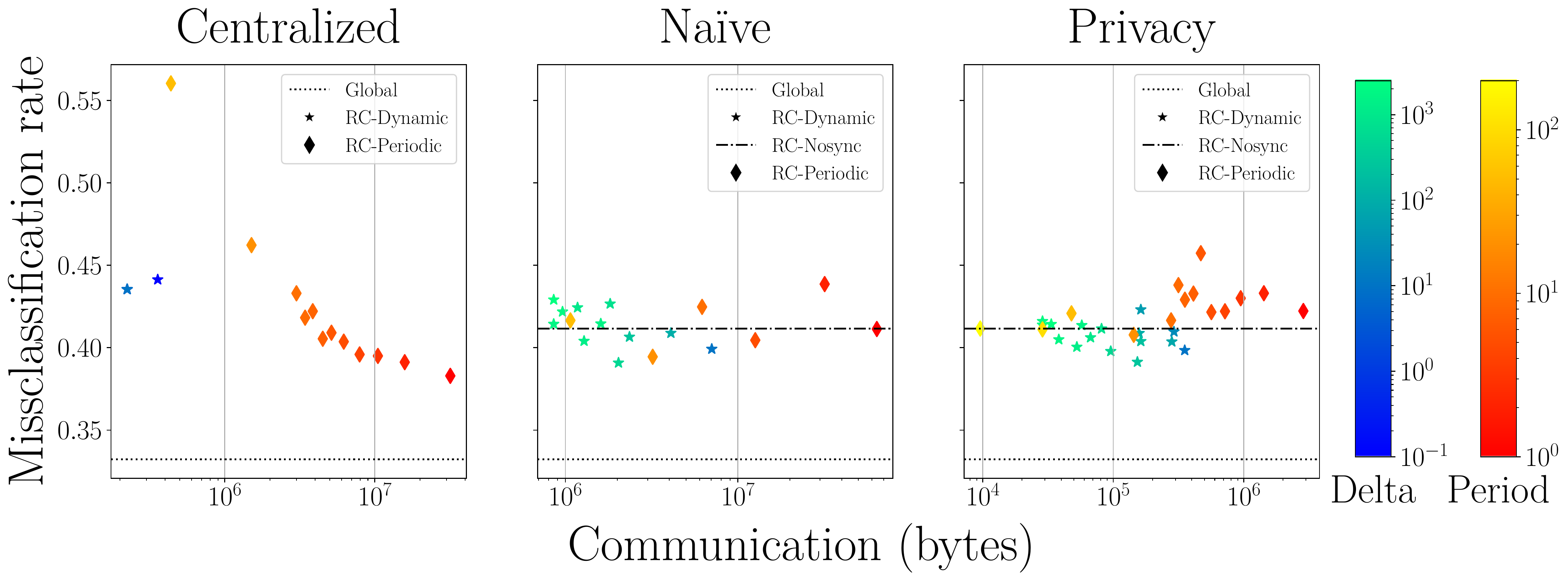}
    \caption{Missclassification rate vs Communication (\texttt{Covertype})}
    \label{fig:rc_covertype}
\end{figure}

\begin{figure}[t]
    \centering
    \includegraphics[width=\linewidth]{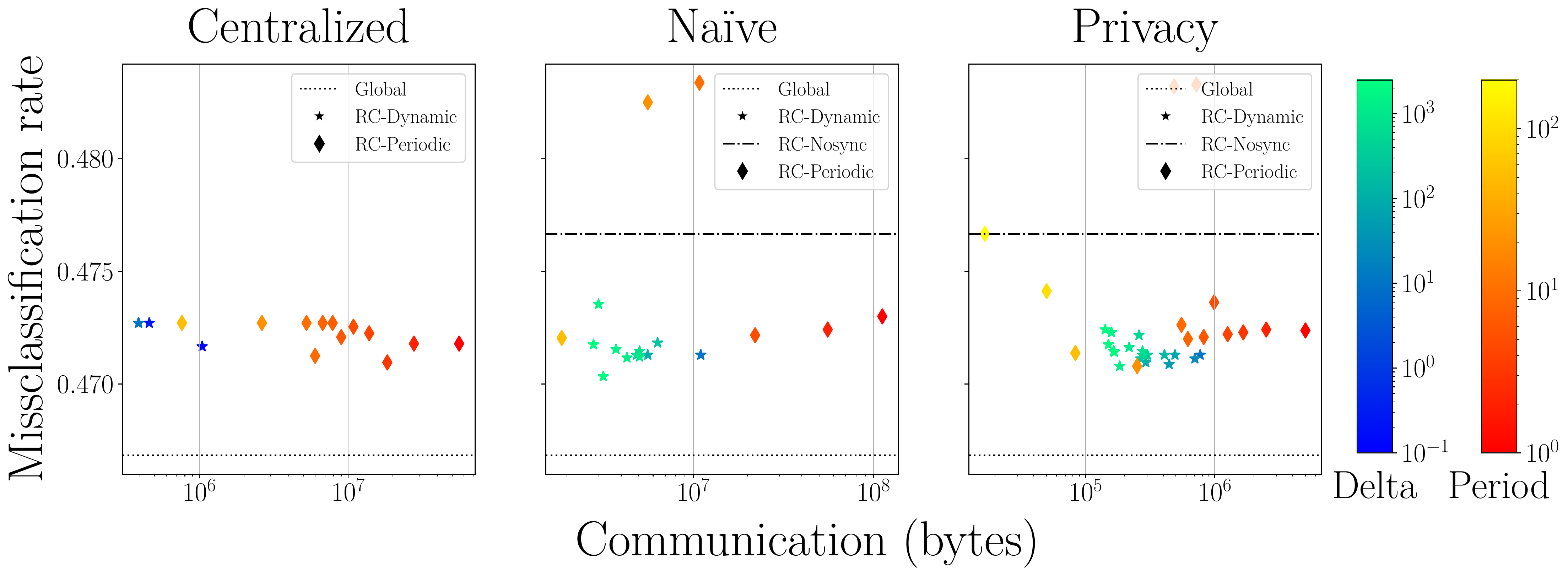}
    \caption{Missclassification rate vs Communication (\texttt{DOTA2})}
    \label{fig:rc_dota2}
\end{figure}

\begin{figure}[t]
    \centering
    \includegraphics[width=\linewidth]{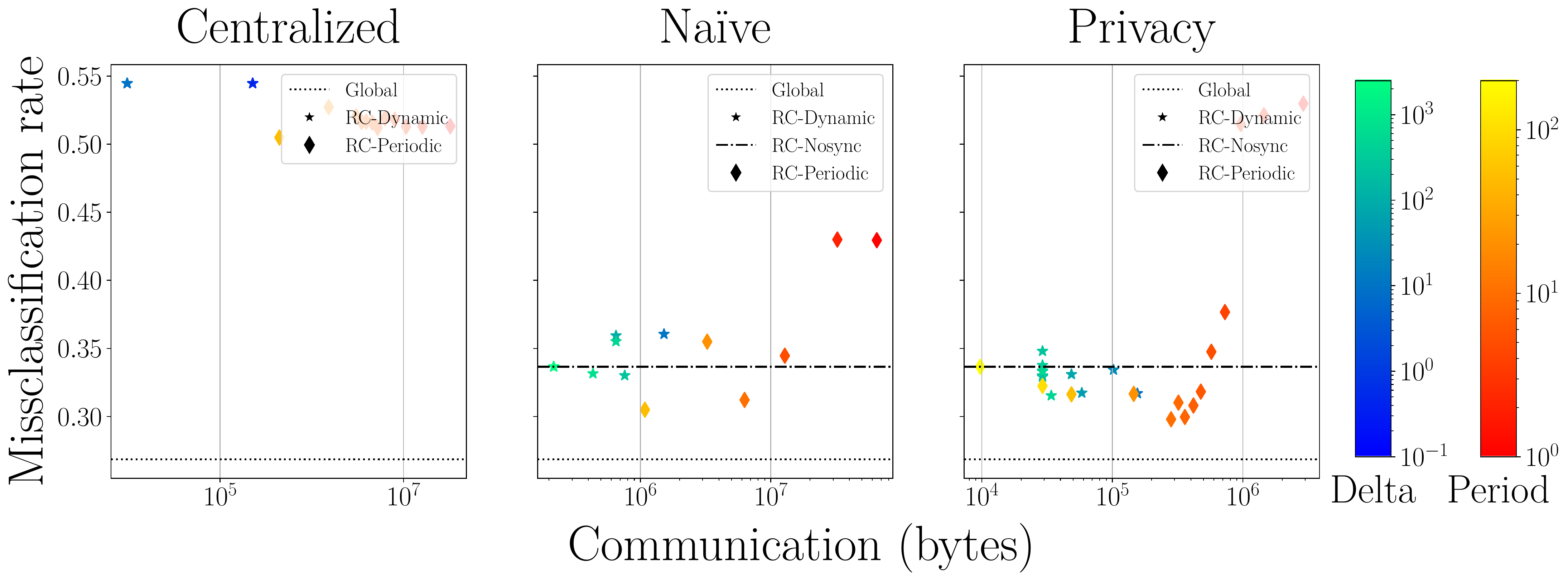}
    \caption{Missclassification rate vs Communication (\texttt{SUSY})}
    \label{fig:rc_susy}
\end{figure}

We have plotted the resource constrained performance of dynamic versus periodic updates for the three datasets and aggregation-mechanisms in Fig. \ref{fig:rc_covertype}, \ref{fig:rc_dota2} and \ref{fig:rc_susy}. The y-axis shows the error and the x-axis shows the communication consumption in bytes. The baseline of no synchronization is shown by a dashed-dotted line. We further include the performance of the global baseline as dotted line. The periodic and dynamic approaches are displayed with different markers, while different periods and deltas are displayed with varying color choices. Some periodically updated models are worse, but most models are superior to the base line. As expected, for \texttt{DOTA2} and \texttt{COVERTYPE}, we see that the dynamic update requires less communication in both, the private and the public setting (little stars in the lower left corner). In the numerical dataset \texttt{SUSY}, the dynamic update uses less communication resources, but has less accuracy than the periodic one in the public setting. We also see that the quality of the periodically updated models varies much more than the one of the dynamically updated ones. Although for the dataset with only numerical attributes, \texttt{SUSY}, the results are not clearly favoring the dynamic update, overall, the answer to the first question leans towards the dynamic protocol. Furthermore, we notice the privacy-preserving aggregation retains the same predictive quality as the na\"ive approach while reducing the required communication substantially. Thus we focus on the privacy-preserving protocol for the next question.

\begin{figure}[t]
    \centering
    \includegraphics[width=\linewidth]{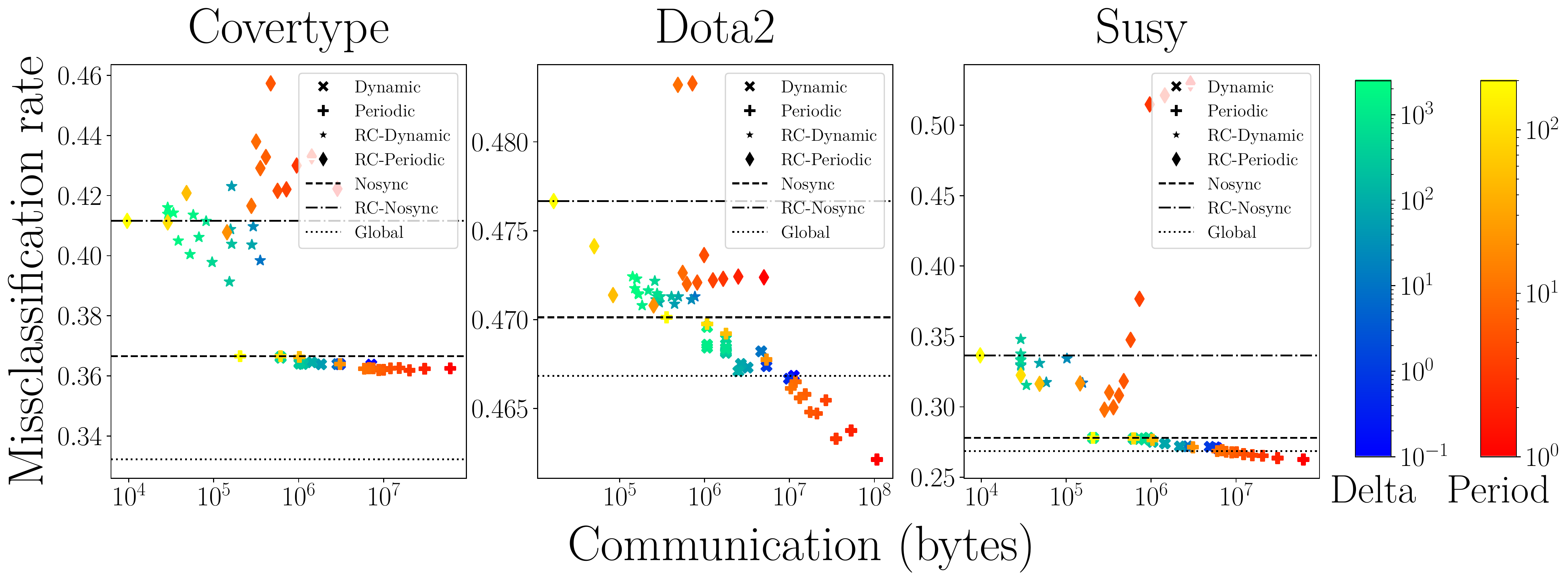}
    \caption{Privacy-Preserving-Resource-Constrained-Averaging}
    \label{fig:Privates}
\end{figure}

% Fig. \ref{fig:Privates} shows the results of the resource-constrained as well as the regular models using the privacy-preserving protocol across the three datasets. 
%For the resource-constrained models,
% Fig . \ref{fig:Privates} shows the 

%TODO: I think the baseline senctences have to be adopted
Fig. \ref{fig:Privates} shows the privacy-preserving scheme for the three datasets comparing the resource-constrained and the normal graphical models. Again, the y-axis  shows the error and the x-axis shows the communication consumption. The dotted baseline shows the performance of the global model, while the dotted baseline stands for not synchronizing normal models and the dashed-dotted line for not synchronizing resource-constrained models. We have varied the frequency of updates for the normal models as well as for the resource constrained ones using the periodic or the dynamic scheme. The parameters for the frequency of update show not surprisingly that more frequent synchronization leads to better model quality which introduces more communication costs as can be seen below the benchmark spreading towards the right. Many of the resource constrained private models outperform the RC baseline. The variance is due to the frequency of updates. The most accurate model is the not resource constrained one in the lower right corner. It uses by far the most communication. We have varied the parameters and see that actually most models of the resource-constrained dynamic updates (stars) outperform the resource-constrained baseline and approach the regular one. However, we must admit that there are models of the resource-constrained periodically and dynamically updated schemes that are worse than the baseline. These have a low frequency of update in common. In the plots, we easily recognize the parameter choice, which balances quality (low) and communication (left). Though, in all settings, we can save communication cost of 1-3 magnitudes while dropping a few percents of classification performance. This is a natural trade-off we encounter in resource-constrained machine learning methods.

\subsection{Energy Savings}
In this section we provide a rough estimate on the amount of energy that could be saved using resource-constraint distributed learning instead of centralizing data. For that, we compare the energy required to centralize all data and train a model to the energy required for locally training models and averaging them. In this simplified scenario, we do not assume that the centrally computed model needs to be transferred regularly to the local learners. 

To compute the energy required for communication, we assume the data is transmitted over 3G, requiring around $2.9$ kWh/GB~\citep{pihkola2018evaluating}, i.e., $\sigma=0.0029$ Wh/GB.  Furthermore, we assume the central computation is performed on a $p_c=100$ Watt processor and one of the parallel low-energy processors consume $p_p=1$ Watt. Since these low-energy processors (e.g., FPGAs) are specialized hardware, the execution time for aggregating data or training a models is usually shorter than on CPUs~\citep{asano2009performance}. However, for simplicity we assume similar runtimes. 

Let $c_c\in\R$ denote the amount of communication in GB required by the central approach and $c_p\in\R$ the amount for the parallel one. Let $a,t\in\R$ denote the time required for aggregating $N\in\N$ data points, respectively training a model. With $m\in\N$ local learners, the energy consumed by the central approach then is 
\[
e_c = \underbrace{(mNa + t)p_c}_{\text{central training}} + c_c\sigma
\]
and the energy for the parallel approach is
\[
e_p = m\underbrace{(Na + t)p_p}_{\text{local training}} + \underbrace{map_p}_{\text{model aggr.}} +  c_p\sigma
\]
As reference, we use the empirical results on the SUSY dataset, where the centralized approach achieves an accuracy of $0.73$ using $6200640B$ of communication and dynamic averaging achieves a comparable accuracy of $0.69$ with $34020B$ of communication.
With $a=10^{-12}$ (roughly the cost of an integer operation on a 1GHz processor) and $t= 10^{-10}$ (roughly $100$ integer operations on a 1GHz processor), $m=16$ and $N=100$ the rough estimates are $e_c=0.22\mu Wh$ and $e_p=0.0033\mu Wh$, i.e., the centralized approach requires more than $67$ times more energy than the parallel one. Note that this estimate is conservative since it does not take into account the energy required in the centralized approach for running an operating system or powering additional components of a computer.

\TwoFig{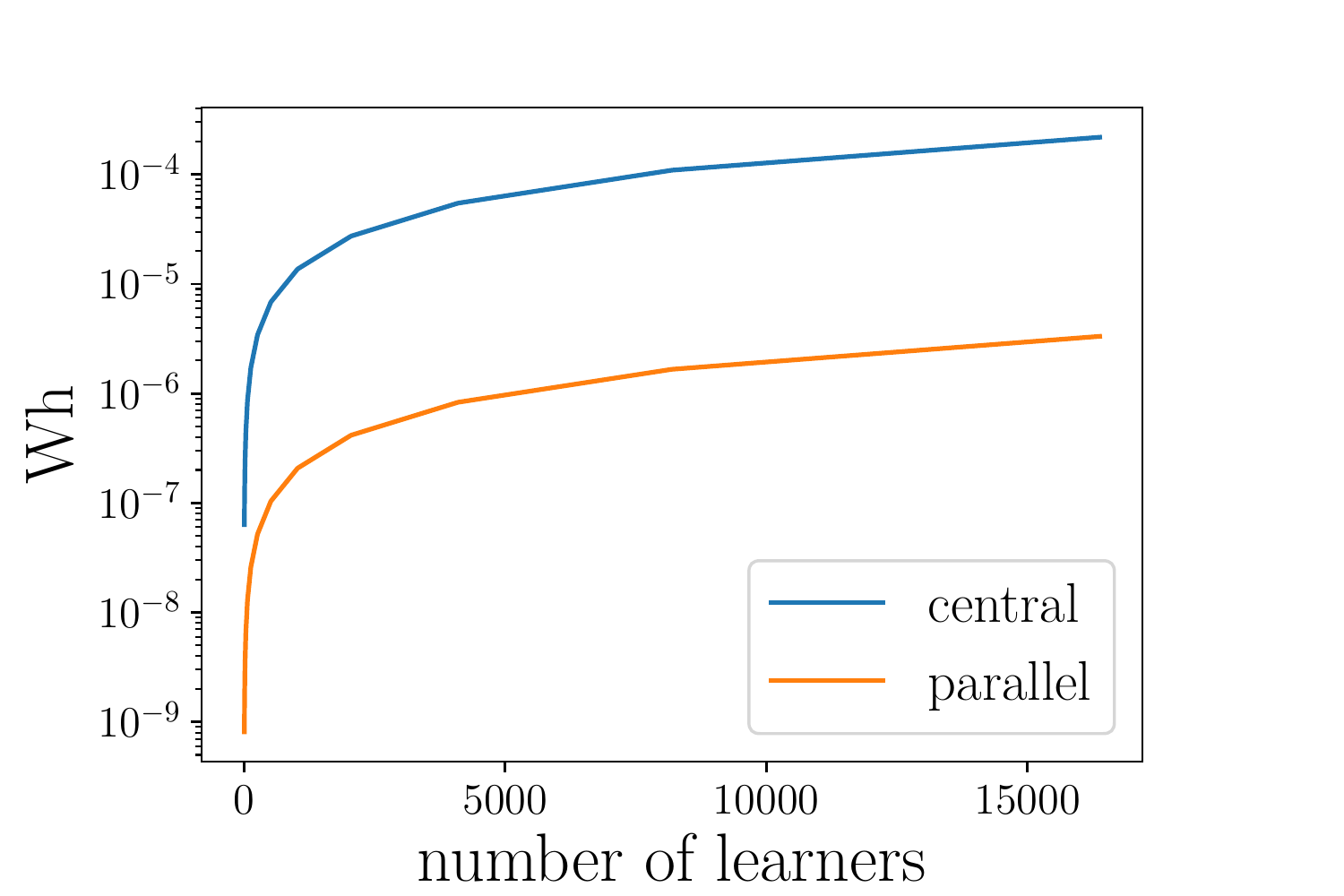} % image 1
     {Scaling behavior of the estimated energy consumption with the number of learners.} % caption 1
     {fig:energyVsNumberLearners} % label 1
     {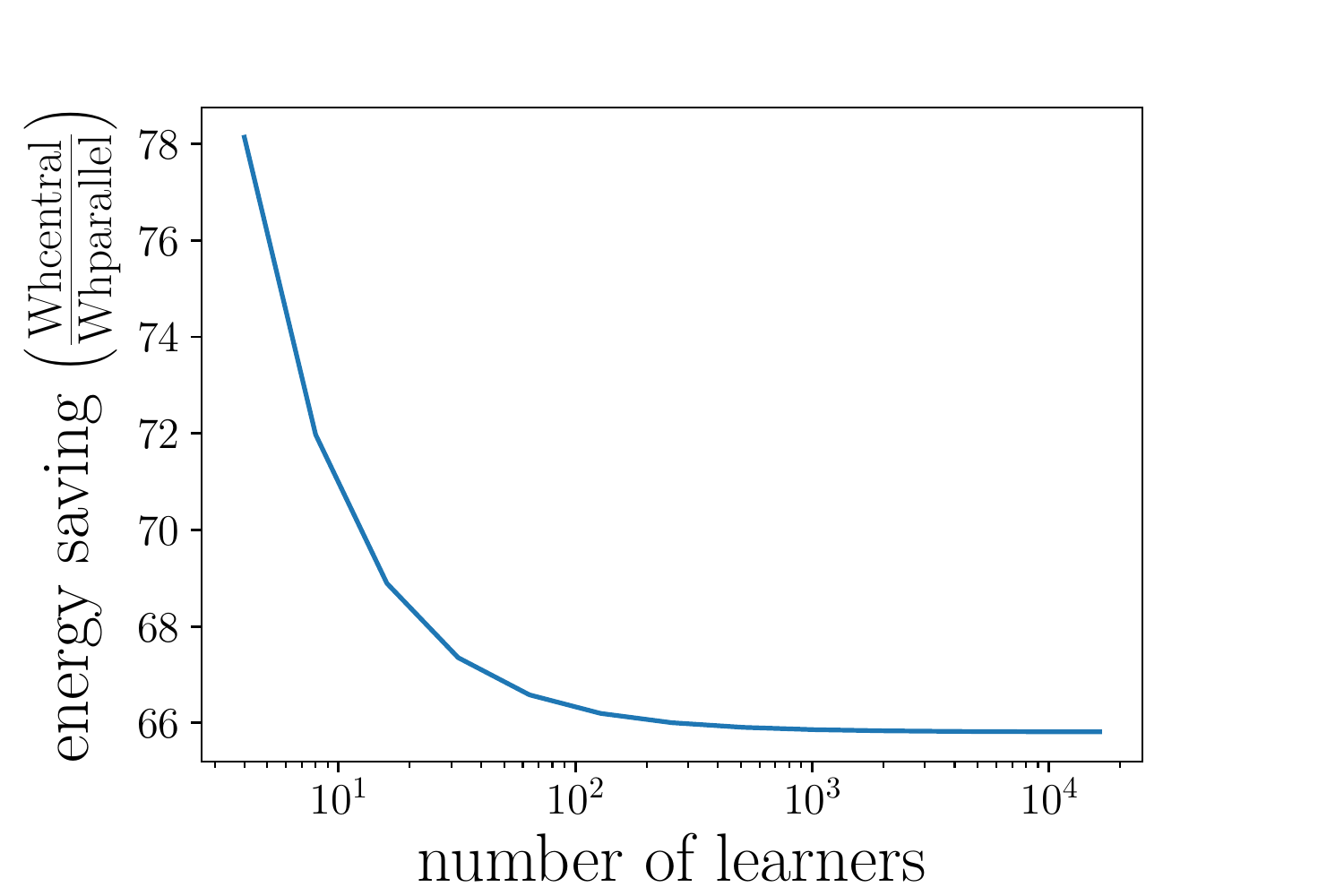} % image 2
     {Scaling behavior of the energy saving's ratio with the number of learners.} % caption 2
     {fig:energySavingsVsNumberLearners} % label 2

% \begin{figure}[tb] %[tb]
% \begin{minipage}{0.48\textwidth} % \textwidth
%   \centering
%   \includegraphics[width=\linewidth]{pics/energySaving/energyPerLearner.pdf} % , width=\linewidth
%   \caption{def}% Scaling behavior of the estimated energy consumption with the number of learners.
%   \label{fig:energyVsNumberLearners}
% \end{minipage}%
% \quad
% \begin{minipage}{0.48\textwidth} % \textwidth
%   \centering
%   \includegraphics[width=\linewidth]{pics/energySaving/energySaving.pdf} % , width=\linewidth
%   \caption{abc} % Scaling behavior of the energy saving's ratio with the number of learners.
%   \label{fig:energySavingsVsNumberLearners}
% \end{minipage}
% \minipage{0.48\textwidth}
%   \centering
%   \includegraphics[width=\linewidth]{pics/energySaving/energyPerLearner.pdf} % , width=\linewidth
%   \caption{Scaling behavior of the estimated energy consumption with the number of learners.}
%   \label{fig:energyVsNumberLearners}
% \endminipage\hfill
% \quad
% \minipage{0.48\textwidth}
%   \centering
%   \includegraphics[width=\linewidth]{pics/energySaving/energySaving.pdf} %, width=\linewidth
%   \caption{Scaling behavior of the energy saving's ratio with the number of learners.}
%   \label{fig:energySavingsVsNumberLearners}
% \endminipage
% \end{figure}
To analyze the potential scaling behavior we make one more simplifying assumptions: We assume that the ratio of central to parallel communication remains constant with the number of learners. This assumption is not very realistic, but the actual scaling behavior depends on the underlying learning problem. Increasing the number of learners typically leads to an even more favorable ratio for simple learning problems, whereas for hard learning problems it can be less favorable~\citep{kamp2019black}. Under this assumption, we show the estimated energy consumption in Figure~\ref{fig:energyVsNumberLearners}. It shows that the energy consumption of the parallel approach remains substantially lower for larger amounts of learners. We show the relative reduction in energy consumption (i.e., $e_c/e_p$) in Figure~\ref{fig:energySavingsVsNumberLearners}. It shows that the relative reduction decreases with the number of learners, but remains above $65$ even for large numbers of learners.

%% file: conclusion.tex
\section{Discussion}
As the empirical evaluation has shown, using only integer operations allows to successfully train models distributedly. However, when using the integer average the relationship between communication and model quality is not as clear as for normal averaging, where more communication reliably leads to higher model quality. The results for resource-constrained averaging indicate that the errors through rounding the average lead to less predictable behavior. The results on SUSY furthermore indicate that too much communication can even be harmful. A possible explanation is that when averaging very often, small changes of local models will get leveled out by the rounding. Thus, the effect of local training is nullified, delaying the overall training process. 

This rounding effect might vanish when using larger numbers of learners (in the experiments, we only used $m=16$ learners). At the same time, this can influence the communication reduction. When computing the energy savings, we assumed the reduction remains constant with larger numbers of learners. It is conceivable for dynamic averaging that larger numbers of learners lead to an even greater reduction in communication when compared to the centralized approach. However, the effects of rounding might similarly lead to more local violations and ultimately a higher amount of communication. To answer this question and determine whether the approach is useful in practice, it is necessary to further study the effects of rounding the average, both empirically and theoretically.

\section{Conclusion and Future work}

In this paper, we proposed a new resource-constrained averaging operator, which can be evaluated on ultra-low-power hardware using only integer operations. Besides, we have shown, that the same applies to the evaluation of the local conditions, which consecutively allows performing every step in this distributed learning setting in a resource-constrained fashion. Furthermore, we have shown that the excess loss of using the dynamic averaging protocol over the periodic protocol is bounded. In our experiments we verified that using resource-constrained averaging of integer exponential family models, we reach a similar performance in terms of prediction quality compared to regular exponential family models with access to the non-restricted parameter space, while reducing network requirements substantially. Besides, we do not only save energy by reducing network communication, but also by employing these models on specific, cheap available hardware. However, there is still a decrease in predictive quality from using resource-constrained models, so this trade-off has to be taken into account: Are we willing to drop $x\%$ accuracy for the sake of energy or bandwidth savings?

% Sentence with partial aggregation isn't really nice...
\subsubsection{Future work} This work opens many new exciting research questions, e.g., how does the averaging perform if we vary the number of nodes in our learning environment. Does it increase the error or do more nodes provide more information for faster convergence? Furthermore, it would be interesting to investigate the effects of resource-constrained-averaging on the parameter vectors in a controlled environment using artificial datasets with known parameters. Besides, the dynamic averaging protocol might yield better results in terms of communication cost in the presence of more nodes, since the partial synchronization mechanism will be triggered more often. Another way to reduce communication could be the clique-wise transmission of data summaries and/or parameters. This technique could possibly help to overcome the limitations of the fixed graph structure by matching the cliques of the different learners and only averaging their parameters. Also, we saw that using incorrect choices for the synchronization period and/or delta, we receive sub-optimal solutions with low predictive quality. This raises the need for informed methods to choose those hyperparameters, incorporating the parameter space, the number of processed samples as well as the optimization budget. Furthermore, since we only considered plain averaging in this work, we could also try to adopt different aggregation mechanisms, e.g., performance-weighted averages or the Radon machine~\citep{kamp2017effective}, to the resource-constrained setting. Further work could also examine the adaptivity to time-variant distributions as well as the performance / convergence on non i.i.d datasets.

% More complex structure

\subsubsection{Acknowledgement} This  research  has  been  funded  by  the  Federal  Ministry of  Education  and  Research  of  Germany  as  part  of  the  competence  center  for machine learning ML2R (01$\vert$S18038A/B/C).

%% file: main.bbl
\begin{thebibliography}{20}
\providecommand{\natexlab}[1]{#1}
\providecommand{\url}[1]{\texttt{#1}}
\providecommand{\urlprefix}{}

\bibitem[{Administration(2018)}]{american2018how}
Administration, U.E.I.: How much electricity does a typical nuclear power plant
  generate?
\newblock
  \url{https://www.americangeosciences.org/critical-issues/faq/how-much-electricity-does-typical-nuclear-power-plant-generate}
  (2018), accessed: 02.12.2019

\bibitem[{Asano et~al.(2009)Asano, Maruyama, and
  Yamaguchi}]{asano2009performance}
Asano, S., Maruyama, T., Yamaguchi, Y.: Performance comparison of fpga, gpu and
  cpu in image processing.
\newblock In: 2009 FPL. pp. 126--131. IEEE (2009)

\bibitem[{Chow and Liu(2006)}]{Chow/Liu/1968a}
Chow, C., Liu, C.: Approximating discrete probability distributions with
  dependence trees.
\newblock IEEE Trans. Inf. Theor. 14(3), 462–467 (Sep 2006)

\bibitem[{Dua and Graff(2017)}]{Dua:2019}
Dua, D., Graff, C.: {UCI} machine learning repository (2017),
  \urlprefix\url{http://archive.ics.uci.edu/ml}

\bibitem[{Hammersley and Clifford(1971)}]{HammersleyClifford1971theorem}
Hammersley, J.M., Clifford, P.E.: Markov random fields on finite graphs and
  lattices.
\newblock {\em Unpublished manuscript}  (1971)

\bibitem[{Kamp(2019)}]{kamp2019black}
Kamp, M.: Black-Box Parallelization for Machine Learning.
\newblock Ph.D. thesis, Rheinische Friedrich-Wilhelms-Universit\"{a}t Bonn
  (2019)

\bibitem[{Kamp et~al.(2018)Kamp, Adilova, Sicking, H{\"u}ger, Schlicht, Wirtz,
  and Wrobel}]{kamp2018efficient}
Kamp, M., Adilova, L., Sicking, J., H{\"u}ger, F., Schlicht, P., Wirtz, T.,
  Wrobel, S.: Efficient decentralized deep learning by dynamic model averaging.
\newblock In: ECML-PKDD. pp. 393--409. Springer (2018)

\bibitem[{Kamp et~al.(2014)Kamp, Boley, Keren, Schuster, and
  Sharfman}]{kamp2014communication}
Kamp, M., Boley, M., Keren, D., Schuster, A., Sharfman, I.:
  Communication-efficient distributed online prediction by dynamic model
  synchronization.
\newblock In: ECML-PKDD. pp. 623--639. Springer (2014)

\bibitem[{Kamp et~al.(2017)Kamp, Boley, Missura, and
  G\"{a}rtner}]{kamp2017effective}
Kamp, M., Boley, M., Missura, O., G\"{a}rtner, T.: Effective parallelisation
  for machine learning.
\newblock In: Advances in Neural Information Processing Systems 30, pp.
  6477--6488. Curran Associates, Inc. (2017)

\bibitem[{Kamp et~al.(2016)Kamp, Bothe, Boley, and
  Mock}]{kamp2016communication}
Kamp, M., Bothe, S., Boley, M., Mock, M.: Communication-efficient distributed
  online learning with kernels.
\newblock In: Joint European Conference on Machine Learning and Knowledge
  Discovery in Databases. pp. 805--819. Springer (2016)

\bibitem[{Kemp(2019)}]{hootsuite2019digital}
Kemp, S.: Digital 2019 global digital overview.
\newblock
  \url{https://datareportal.com/reports/digital-2019-global-digital-overview}
  (2019), accessed: 02.12.2019

\bibitem[{Lim et~al.(2019)Lim, Luong, Hoang, Jiao, Liang, Yang, Niyato, and
  Miao}]{DBLP:journals/corr/abs-1909-11875}
Lim, W.Y.B., Luong, N.C., Hoang, D.T., Jiao, Y., Liang, Y., Yang, Q., Niyato,
  D., Miao, C.: Federated learning in mobile edge networks: {A} comprehensive
  survey.
\newblock CoRR abs/1909.11875 (2019),
  \urlprefix\url{http://arxiv.org/abs/1909.11875}

\bibitem[{McMahan et~al.(2017)McMahan, Moore, Ramage, Hampson, and
  y~Arcas}]{mcmahan2017communication}
McMahan, B., Moore, E., Ramage, D., Hampson, S., y~Arcas, B.A.:
  Communication-efficient learning of deep networks from decentralized data.
\newblock In: Artificial Intelligence and Statistics. pp. 1273--1282 (2017)

\bibitem[{Mohan and Kangasharju(2016)}]{mohan2016edge}
Mohan, N., Kangasharju, J.: Edge-fog cloud: A distributed cloud for internet of
  things computations.
\newblock In: 2016 Cloudification of the Internet of Things (CIoT). pp. 1--6.
  IEEE (2016)

\bibitem[{Piatkowski(2019)}]{Piatkowski/2019a}
Piatkowski, N.: Distributed generative modelling with sub-linear communication
  overhead.
\newblock In: Kamp, M., Paurat, D., Krishnamurthy, Y. (eds.) Decentralized
  Machine Learning at the Edge. Springer (2019)

\bibitem[{Piatkowski et~al.(2016)Piatkowski, Lee, and
  Morik}]{DBLP:journals/ijon/Piatkowski0M16}
Piatkowski, N., Lee, S., Morik, K.: Integer undirected graphical models for
  resource-constrained systems.
\newblock Neurocomputing 173, 9--23 (2016)

\bibitem[{Piatkowski(2018)}]{piatkowski2018exponential}
Piatkowski, N.P.: Exponential families on resource-constrained systems.
\newblock Ph.D. thesis, TU Dortmund (2018)

\bibitem[{Pihkola et~al.(2018)Pihkola, Hongisto, Apilo, and
  Lasanen}]{pihkola2018evaluating}
Pihkola, H., Hongisto, M., Apilo, O., Lasanen, M.: Evaluating the energy
  consumption of mobile data transfer—from technology development to consumer
  behaviour and life cycle thinking.
\newblock Sustainability 10(7), 2494 (2018)

\bibitem[{Shi et~al.(2016)Shi, Cao, Zhang, Li, and Xu}]{shi2016edge}
Shi, W., Cao, J., Zhang, Q., Li, Y., Xu, L.: Edge computing: Vision and
  challenges.
\newblock IEEE Internet of Things Journal 3(5), 637--646 (2016)

\bibitem[{Wainwright and Jordan(2008)}]{Wainwright/etal/2008a}
Wainwright, M.J., Jordan, M.I.: Graphical models, exponential families, and
  variational inference.
\newblock Found. Trends Mach. Learn. 1(1-2) (2008)

\end{thebibliography}
